\documentclass{article}
\usepackage{amssymb}
\usepackage{xcolor}
\usepackage{graphicx}
\usepackage[implicit=false]{hyperref}

\long\def\comment#1{}

\newtheorem{assumption}{Assumption}
\newtheorem{theorem}{Theorem}

\newtheorem{lemma}[theorem]{Lemma}

\newcommand{\Thm}[1]{Theorem~\ref{#1}}
\newcommand{\Lem}[1]{Lemma~\ref{#1}}

\newcommand{\Ass}[1]{Assumption~\ref{#1}}

\newcommand{\Fig}[1]{Figure~\ref{#1}}

\newcommand{\Eq}[1]{Equation~\ref{#1}}
\newcommand{\Eqs}[1]{Equations~\ref{#1}}
\newcommand{\qed}{\mbox{$\Box$}}
\newcommand{\Sec}[1]{Section~\ref{#1}}

\newcommand{\Db}{D}

\newcommand{\Npij}{N'_{ij}}
\newcommand{\Npijk}{N'_{ijk}}

\newcommand{\Bs}{B_s}
\newcommand{\Bp}{B_p}

\newcommand{\Bsc}{B_{sc}}

\newcommand{\Bszero}{B_{s0}}
\newcommand{\Bsone}{B_{s1}}
\newcommand{\Bstwo}{B_{s2}}

\newcommand{\hBs}{B^h_s}

\newcommand{\hBsc}{B^h_{sc}}

\newcommand{\hBszero}{B^h_{s0}}
\newcommand{\hBsone}{B^h_{s1}}
\newcommand{\hBstwo}{B^h_{s2}}

\newcommand{\hBxtoy}{B^h_{x \rightarrow y}}
\newcommand{\hBytox}{B^h_{x \leftarrow y}}
\newcommand{\hBxy}{B^h_{xy}}

\newcommand{\dXdY}[2]{\partial {#1} / \partial {#2}}

\newcommand{\xtoy}{x \rightarrow y}
\newcommand{\ytox}{y \rightarrow x}

\newcommand{\Th}[1]{\Theta_{#1}}

\newcommand{\TBs}{\Theta_{Bs}}

\newcommand{\TBsc}{\Theta_{Bsc}}

\newcommand{\ta}[1]{\theta_{#1}}

\newcommand{\m}{\vec{m}} 
\newcommand{\vecv}{\vec{v}} 
\newcommand{\vecb}{\vec{b}} 
\newcommand{\vecbi}{\vec{b_i}} 
 
\newcommand{\matW}{W} 
\newcommand{\matS}{S} 
\newcommand{\matM}{M} 
\newcommand{\matSigma}{\Sigma} 
\newcommand{\matTau}{T} 
\newcommand{\x}{\vec{x}}

\newcommand{\xvecbar}{\overline{\vec{x}}} 
\newcommand{\vecx}{\vec{x}} 
 
\newcommand{\vecm}{\vec{m}} 
\newcommand{\vecmu}{\vec{\mu}} 
 
\newcommand{\vecbji}{b_{ji}}

\newcommand{\matB}{B} 
\newcommand{\tr}{\mbox{\it tr}} 
\newcommand{\Npmu}{N'_{\vec{\mu}}}
\newcommand{\Nptau}{N'_{\matTau}}

\title{Learning Bayesian Networks:\\ A Unification for Discrete and 
Gaussian Domains}

\comment{
\author{
David Heckerman and Dan Geiger\thanks{\ Author's primary affiliation:
Computer Science Department, Technion, Haifa 32000, Israel.}\\
\\
Microsoft Research, Bldg 9S/1\\
Redmond 98052-6399, WA\\
heckerma@hotmail.com,geiger02@gmail.com}
}

\author{
David Heckerman\\
heckerma@hotmail.com\\
\\
Dan Geiger\\ 
geiger02@gmail.com}

\date{August 1995, Revised June 2021}

\begin{document}

\maketitle

\begin{abstract}

\noindent 
We examine Bayesian methods for learning Bayesian networks from a
combination of prior knowledge and statistical data.  In particular,
we unify the approaches we presented at last year's conference for
discrete and Gaussian domains.  We derive a general Bayesian scoring
metric, appropriate for both domains.  We then use this metric in
combination with well-known statistical facts about the Dirichlet and
normal--Wishart distributions to derive our metrics for discrete and
Gaussian domains.

\bigskip

\noindent
Corrections to the original text in \textcolor{red}{red} are taken
from J. Kuipers, G. Moffa, and D. Heckerman, Addendum on the scoring
of Gaussian directed acyclic graphical models. {\em Annals of Statistics} 
42, 1689-1691, Aug 2014. 
Other updates to the original are in \textcolor{blue}{blue}.

\end{abstract}

\section{Introduction} \label{sec:intro}

At last year's conference, we presented approaches for learning
Bayesian networks from a combination of prior knowledge and
statistical data.  These approaches were presented in two papers: one
addressing domains containing only discrete variables (Heckerman et
al., 1994)\nocite{HGC94uai}, and the other addressing domains
containing continuous variables related by an unknown
multivariate-Gaussian distribution (Geiger and Heckerman,
1994).\nocite{GH94uai} Unfortunately, these presentations were
substantially different, making the parallels between the two methods
difficult to appreciate.  In this paper, we unify the two approaches.
In particular, we abstract our previous assumptions of likelihood
equivalence, parameter modularity, and parameter independence
such that they are appropriate for discrete and Gaussian domains (as
well as other domains).  Using these assumptions, we derive a
domain-independent Bayesian scoring metric.  We then use this general
metric in combination with well-known statistical facts about the
Dirichlet and normal--Wishart distributions to derive our metrics for
discrete and Gaussian domains.  In addition, we provide simple proofs
that these assumptions are consistent for both domains.

Throughout this discussion, we consider a domain $U$ of $n$ variables
$x_1,\ldots,x_n$.  Each variable may be discrete---having a finite or
countable number of states---or continuous.  We use lower-case letters
to refer to variables and upper-case letters to refer to sets of
variables.  We write $x_i=k$ to denote that variable $x_i$ is in state
$k$.  When we observe the state for every variable in set $X$, we call
this set of observations a {\em state} of $X$; and we write
$X=k_X$ as a shorthand for the observations $x_i=k_i, x_i \in
X$.  The {\em joint space} of $U$ is the set of all states of $U$.
We use $p(X=k_X|Y=k_Y,\xi)$ to denote the generalized
probability density that $X=k_X$ given $Y=k_Y$ for a
person with current state of information $\xi$ [DeGroot, 1970,
p.\ 19].\nocite{DeGroot70} We use $p(X|Y,\xi)$ to denote the
generalized probability density function (gpdf) for $X$, given all
possible observations of $Y$.  The {\em joint gpdf} over $U$ is the
gpdf for $U$.

We use $\Bs$ to denote the structure of a Bayesian network, and
$\Pi_i$ to denote the parents of $x_i$ in a given network.  We assume
the reader is familiar with Bayesian networks for the case where all
variables in $U$ are discrete.  Here, we describe a Bayesian-network
representation for continuous variables.  In particular, consider the
special case where all the variables in $U$ are continuous and the
joint probability density function for $U$ is a multivariate
(nonsingular) normal distribution.  In this case, to be in line with
more standard notation, we use $\x$ to denote the set of variables
$U$.  We have
\begin{eqnarray}  \label{eq:normal}      
p(\x|\xi) & = & n(\vecmu,\matSigma^{-1}) \\
  & \equiv &
  (2 \pi)^{-n/2} |\matSigma|^{-1/2}
  e^{-1/2 (\x - \vecmu)' \matSigma ^{-1} (\x - \vecmu)} \nonumber
\end{eqnarray} 
where $\vecmu$ is an $n$-dimensional mean vector, and $\matSigma =
(\sigma_{ij})$ is an $n \times n$ covariance matrix, which must be
both symmetric and positive definite.  Both $\vecmu$ and $\matSigma$
are implicitly functions of $\xi$.  We shall find it convenient to
refer to the {\em precision matrix} $\matW = \matSigma^{-1}$, whose
elements are denoted by $w_{ij}$.
 
This joint density function can be written as a product of conditional
density functions each being a normal distribution.  Namely,
\begin{equation} \label{eq:prod-n} 
p(\x|\xi)= \prod_{i=1}^n 
  p(x_i | x_1, \ldots,x_{i-1},\xi) 
\end{equation} 
\begin{equation} \label{eq:cond-normal} 
p(x_i | x_1, \ldots,x_{i-1}, \xi) = 
   n(\mu_i + \sum_{j=1}^{i-1} \vecbji (x_j - \mu_j), 1/v_i) 
\end{equation} 
where $\mu_i$ is the unconditional mean of $x_i$ (i.e., the $i$th
component of $\vecmu$), $v_i$ is the conditional variance of $x_i$
given values for $x_1, \ldots, x_{i-1}$, and $b_{ji}$ is a linear
coefficient reflecting the strength of the relationship between $x_j$
and $x_i$ (e.g., DeGroot, p.\ 55).

Thus, we may interpret a multivariate-normal distribution as a
Bayesian network, where there is no arc from $x_j$ to $x_i$ whenever
$b_{ji} = 0$, $j<i$.  Conversely, from a Bayesian network with
conditional distributions satisfying \Eq{eq:cond-normal}, we may
construct a multivariate-normal distribution.  We call this special
form of a Bayesian network a {\em Gaussian network.}  The name is adopted
from Shachter and Kenley (1989) \nocite{Shachter89b} who first
described Gaussian influence diagrams.  We note that, in practice, it
is typically easier to assess a Gaussian network than it is to assess
directly a symmetric positive-definite precision matrix.

The transformations between $\vecv = \{v_1,\ldots,v_n\}$ and $\matB
\equiv \{b_{ji} \mid j < i \}$ of a given Gaussian network $G$
and the precision matrix $\matW$ of the normal distribution represented
by $G$ are well known.  In this paper, we need only the transformation
from $\matW$ to $\{\vecv, \matB\}$.  We use the following recursive
form given by Shachter and Kenley (1989).  Let $W(i)$ denote the $i
\times i$ upper left submatrix of $\matW$, $\vecb_i$ denote the column
vector $(b_{1i}, \ldots, b_{i-1,i})$, and $\vecb'_i$ denote the
transpose of $\vecb_i$.  Then, for $i>1$, we have
\begin{equation} 
\label{eq:shachter} 
\matW(i+1) = \left( \begin{array}{cc} 
\matW(i) + \frac{\vecb_{i+1} \vecb'_{i+1}}{v_{i+1}} & 
-\frac{\vecb_{i+1}}{v_{i+1}}   \\ 
-\frac{\vecb'_{i+1}}{v_{i+1}} &  \frac{1}{v_{i+1}} 
\end{array} \right) 
\end{equation} 
and $\matW(1)= \frac{1}{v_1}$. 

Although \Eq{eq:cond-normal} is useful for the assessment
of a Gaussian network, we shall sometimes find it convenient to
write
\begin{equation} \label{eq:cond-normal-m} 
p(x_i | x_1, \ldots,x_{i-1}, \xi) = 
   n(\m_i + \sum_{j=1}^{i-1} \vecbji x_j, 1/v_i) 
\end{equation} 
where $m_i,i=1,\ldots,n$ is defined by
\begin{equation} \label{eq:mi}
m_i = \mu_i - \sum_{j=1}^{i-1} b_{ji} \mu_j
\end{equation}
Note that $m_i$ is the mean of $x_i$ when all of
$x_i$'s parents are equal to zero.
 
As an example, given the three-node network structure $x_1 \rightarrow x_3
\leftarrow x_2$, 
we have $b_{12}=0$, $x_1 = n(m_1, 1/v_1), x_2 = n(m_2, 1/v_2),$ and
$x_3 = n(m_3 + b_{13} (x_1-m_1) + b_{23} (x_2-m_2), 1/v_3)$.  Also,
the precision matrix corresponding to this network structure is given
by
\begin{equation} 
\label{eq:covmat} 
W =  
\left( \begin{array}{ccc} 
\frac{1}{v_1} + \frac{b_{13}^2}{v_3} &  \frac{b_{13}b_{23}}{v_3} &   
  -\frac{b_{13}}{v_3} \\ 
\frac{b_{13}b_{23}}{v_3} & \frac{1}{v_2} + \frac{b_{23}^2}{v_3} &    
  -\frac{b_{23}}{v_3} \\ 
-\frac{b_{13}}{v_3} & -\frac{b_{23}}{v_3} & \frac{1}{v_3} 
\end{array} \right) 
\end{equation} 
 
Finally, it is important to note that two or more Bayesian-network
structures for a given domain can be {\em equivalent} in the sense
that the structures represent the same set of gpdfs for the domain
(Verma and Pearl, 1990)\nocite{Verma90}.  For example, for the three
variable domain $\{x,y,z\}$, each of the network structures $x
\rightarrow y \rightarrow z$, $x \leftarrow y \rightarrow z$, and $x
\leftarrow y \leftarrow z$ represents the gpdfs where $x$ and
$z$ are conditionally independent of $y$, and are therefore
equivalent.  As another example, a {\em complete network structure} is
one that has no missing edges.  In a domain with $n$ variables, there
are $n!$ complete network structures.  All complete network structures
for a given domain represent the same set of gpdfs---namely, all
possible gpdfs---and are therefore equivalent.  In our proofs to
follow, we require the following characterization of equivalent
networks, proved by Chickering (in this proceedings).\nocite{C95uai}

\begin{theorem}[Chickering, 1995] \label{thm:max}
Let $\Bsone$ and $\Bstwo$ be two Bayesian-network structures, and
$R_{\Bsone,\Bstwo}$ be the set of edges by which $\Bsone$ and $\Bstwo$
differ in directionality.  Then, $\Bsone$ and $\Bstwo$ are equivalent
if and only if there exists a sequence of $|R_{\Bsone,\Bstwo}|$
distinct arc reversals applied to $\Bsone$ with the following properties:
\begin{enumerate}
\item After each reversal, the resulting network structure 
contains no directed cycles and is equivalent to $\Bstwo$

\item After all reversals, the resulting network structure
is identical to $\Bstwo$

\item If $x \rightarrow y$ is the next arc to be reversed
in the current network structure, then $x$ and $y$ have the same
parents in both network structures, with the exception that $x$ is
also a parent of $y$ in $\Bsone$
\end{enumerate}
\end{theorem}

\section{A Bayesian Approach for Learning Bayesian Networks} \label{sec:rs}

Our Bayesian approach for learning Bayesian networks can be understood
as follows.  Suppose we have a domain of variables $\{x_1,\ldots,x_n\}
= U$, and a set of cases $\{C_1,\ldots,C_m\} = \Db$ where each case is
a state of some or of all the variables in $U$.  We sometimes refer
to $\Db$ as a database.  We begin with the following {\em random-sample
assumption}: the database is a random sample from some sample
distribution with unknown parameters $\Th{U}$, and this sample
distribution satisfies the conditional-independence assertions of some
network structure $\Bs$ for $U$.  We define $\hBs$ to be the
hypothesis that the sample distribution can be encoded in $\Bs$.

Now, suppose that we wish to determine the gpdf $p(C|\Db,\xi)$---the
generalized probability density function for a new case $C$, given the
database and our current state of information $\xi$.  Rather than
reason about this distribution directly, we assume that the collection
of hypotheses $\hBs$ corresponding to all network structures for $U$
form a mutually exclusive and collectively exhaustive set\footnote{We
comment on this assumption in the following section.} and compute
\begin{displaymath} \label{eq:ec1}
p(C|\Db,\xi) = \sum_{{\rm all\ } \hBs} p(C|\Db,\hBs,\xi) \cdot p(\hBs|\Db,\xi) 
\end{displaymath}
In practice, it is impossible to sum over all possible network
structures.  Consequently, we attempt to identify a small subset $H$
of network-structure hypotheses that account for a large fraction of
the posterior probability of the hypotheses.
Rewriting the previous equation using the fact that
$p(\hBs|\Db,\xi) = p(\Db,\hBs|\xi) / p(\Db|\xi)$, we obtain
\begin{displaymath} \label{eq:ec3}
p(C|\Db,\xi) \approx c \ \sum_{\hBs \in H} p(C|\Db,\hBs,\xi) 
 \cdot p(\Db,\hBs|\xi)
\end{displaymath}
where $c$ is the normalization constant $1/[\sum_{\hBs \in H}
p(\Db,\hBs|\xi)]$.  From this relation, we see that only the relative
posterior probabilities $p(\Db,\hBs|\xi)$ matter.  Thus, we compute this
relative posterior probability, or alternatively, a {\em Bayes'
factor}---$p(\hBs|\Db,\xi)/p(\hBszero|\Db,\xi)$---where $\Bszero$ is some
reference structure such as the empty graph.  We call methods for
computing these relative posterior probabilities Bayesian scoring metrics.

Extending the Bayesian analysis, we use $\TBs$ to denote the
parameters of the sample distribution encoded in the network structure
$\Bs$ given hypothesis $\hBs$.  That is, the parameters $\TBs$
determine the local gpdfs in $\Bp$.
From the rules of probability, we have
\begin{eqnarray} \label{eq:ec4}
\lefteqn{ p(\Db,\hBs|\xi) = p(\hBs|\xi) } \\
&& \cdot \int p(\TBs|\hBs,\xi) \ p(\Db|\TBs,\hBs,\xi) \ 
  d\TBs  \nonumber
\end{eqnarray}
The assessment of the network-structure priors $p(\hBs|\xi)$ is
treated elsewhere (e.g., Buntine, 1991, and Heckerman et al., 1995).
In the following section, we introduce a set of assumptions that
simplifies the assessment of the network-parameter priors
$p(\TBs|\hBs,\xi)$.  In the remainder of this section, we show how to
compute $p(\Db|\TBs,\hBs,\xi)$.

A method for computing this term follows from our random-sample
assumption.  Namely, given hypothesis $\hBs$, it follows that $\Db$ can
be separated into a set of random samples, where these random samples
are determined by the structure of $\Bs$.  First, let us examine this
decomposition when all the variables in $U$ are discrete.  Let
$\ta{X=k_X|Y=k_Y}$ denote the parameter corresponding to
the probability $p(X=k_X|Y=k_Y,\xi)$, where $X$ and $Y$
are disjoint subsets of $U$.  In addition, let $x_{il}$ and $\Pi_{il}$
denote the variable $x_i$ and the parent set $\Pi_i$ in the $l$th
case, respectively; and let $\Db_l$ denote the first $l-1$ cases in the
database.  Then, given $\hBs$, we know that the observations of $x_i$
in those cases where $\Pi_{il}=k_{\Pi_i}$ is a random sample
with parameters $\Th{x_{il}|\Pi_{il}=k_{\Pi_i}}$.  That is,
\[
p(x_{il}=k_i|x_{1l}=k_1,\ldots,x_{(i-1)l}=k_{i-1},\Db_l,
  \TBs,\hBs,\xi)
\]
\begin{equation} \label{eq:ms}
\ \ \  = \ta{x_{il}=k_i|\Pi_{il}=k_{\Pi_i}} 
\end{equation}
where $k_{\Pi_i}$ is the state of $\Pi_{il}$ consistent
with $\{x_{1l}=k_1,\ldots,x_{(i-1)l}=k_{i-1}\}$.  Using
\Eq{eq:ms}, we can compute $p(\Db|\TBs,\hBs,\xi)$ for any
database $\Db$ and network structure $\Bs$ for discrete domain $U$.

Now consider a domain of continuous variables
$\vecx=\{x_1,\ldots,x_n\}$, and suppose the database $\Db$ is a random
sample from a multivariate-normal distribution with parameters $\Th{U}
= \{\vecmu,\matW\}$.  From our discussion in \Sec{sec:intro}, it
follows that, given hypothesis $\hBs$, each variable $x_i$ is a random
sample from a normal distribution with mean $m_i + \sum_{x_j \in
\Pi_{i}} b_{ji} x_j$ and variance $v_i$.  Thus, with
$\TBs=\{\vecm,\matB,\vecv\}$, we have
\begin{eqnarray} \label{eq:gs}
\lefteqn{ p(x_{il}|x_{1l},\ldots,x_{(i-1)l},\Db_l,
  \TBs,\hBs,\xi) } \nonumber \\
&& = n(m_i + \sum_{x_j \in \Pi_{i}} b_{ji} x_{jl}, 1/v_i)
\end{eqnarray}
Using \Eq{eq:gs}, we can compute $p(\Db|\TBs,\hBs,\xi)$
for any $\Db$ and $\Bs$ in a Gaussian domain.

The generalization of \Eqs{eq:ms} and \ref{eq:gs} is straightforward,
and we state it as our first formal assumption.

\begin{assumption}[Random Sample] \label{ass:rs}
Let $\Db=\{C_1,\ldots,C_m\}$ be a database, and $\Bs$ be a network
structure for $U$ determined by variable ordering $(x_1,\ldots,x_n)$.
Let $\Theta(x_i,\Pi_i)$ denote the parameters of the network
associated with variable $x_i$.  Then, for all variables $x_i \in U$,
\begin{eqnarray} \label{eq:rs}
\lefteqn{ p(x_{il}|x_{1l},\ldots,x_{(i-1)l},\Db_l,\TBs,\hBs,\xi) }
  \nonumber \\
&& =  f(\Theta(x_i,\Pi_i), x_{il}, \Pi_{il})
\end{eqnarray}
where $f$ is some function of the parameters $\Theta(x_i,\Pi_i)$ and the
database entries $x_{il}$ and $\Pi_{il}$.
\end{assumption}

\noindent
In the discrete case, we have $\Theta(x_i,\Pi_i) =
\Th{x_i|\Pi_i}$, and $f(\Theta(x_i,\Pi_i), x_{il}, \Pi_{il}) =
\Th{x_{il}|\Pi_{il}}$.  In the Gaussian case, we have
$\Theta(x_i,\Pi_i) = \{m_i,v_i,\vecb_i\}$, and 
$f(\Theta(x_i,\Pi_i), x_{il}, \Pi_{il}) =
n(m_i + \sum_{x_{j} \in \Pi_{i}} b_{ji} x_{jl}, 1/v_i)$.

\section{Informative Priors} \label{sec:ip}

In this section, we derive a general approach for assessing the
network-parameter priors $p(\TBs|\hBs,\xi)$.  Our derivation is based
on four assumptions that are abstracted from our previous work.

\begin{assumption}[Likelihood Equivalence] \label{ass:le}
Given two network structures $\Bsone$ and $\Bstwo$ such that
$p(\hBsone|\xi)>0$ and $p(\hBstwo|\xi)>0$, if $\Bsone$ and $\Bstwo$
are equivalent, then $p(\Th{U}|\hBsone,\xi) =
p(\Th{U}|\hBstwo,\xi)$.
\end{assumption}

\noindent 
Informally, the assumption states that the observation of a database
does not help to discriminate equivalent network structures.  We note
that an equivalent way to state likelihood equivalence is that
$p(\Db|\hBsone,\xi) = p(\Db|\hBstwo,\xi)$ for all databases $\Db$, whenever
$\Bsone$ and $\Bstwo$ are equivalent.\footnote{We assume this
equivalence is well known, although we have not found a proof in the
literature.}

The motivation for this assumption is different for acausal Bayesian
networks---Bayesian networks that represent only assertions of
conditional independence---and causal Bayesian networks.  For acausal
networks, likelihood equivalence is not an assumption, but rather a
consequence of our definition of $\hBs$.  In particular, recall that
the hypothesis $\hBs$ is true iff the parameters $\Th{U}$ satisfy the
conditional independence assertions of $\Bs$.  Therefore, by
definition of network-structure equivalence, if $\Bsone$ and $\Bstwo$
are equivalent, then $\hBsone=\hBstwo$.\footnote{We note that there is
a flaw with our definition of $\hBs$ for acausal Bayesian networks.
In particular, the definition implies that hypotheses associated with
different network-structure equivalence classes will not be mutually
exclusive.  For example, in the two-binary-variable domain, the
hypotheses $\hBxy$ and $\hBxtoy$ (corresponding to the empty network
structure, and the network structure $\xtoy$, respectively) both
include the possibility $\ta{xy}=\ta{x}\ta{y}$.  This flaw is
potentially troublesome, because mutual exclusivity is important for
our Bayesian interpretation of network learning (\Eq{eq:ec3}).
Nonetheless, because the densities $p(\TBs|\hBs,\xi)$ must be
integrable and hence bounded, the overlap of hypotheses will be of
measure zero, and we may use \Eq{eq:ec3} without modification.  For
example, in our two-binary-variable domain, given the hypothesis
$\hBxtoy$, the probability that $\hBxy$ is true (i.e.,
$\ta{y}=\ta{y|x}$) has measure zero.}  For example, in the domain
$\{x_1,x_2,x_3\}$, the equivalent network structures $x_1
\rightarrow x_2 \rightarrow x_3$ and $x_1 \leftarrow x_2 \leftarrow x_3$
both correspond to the assertion
$\ta{x_1,x_3|x_2}=\ta{x_1|x_2}\ta{x_3|x_2}$.  Consequently, $B^h_{x_1
\rightarrow x_2 \rightarrow x_3} = B^h_{x_1 \leftarrow x_2 \leftarrow
x_3}$.  This property, which we call {\em hypothesis equivalence},
implies likelihood equivalence.  We note that, given hypothesis
equivalence, we should score equivalence classes of network
structures---not individual network structures---when learning acausal
Bayesian networks.

For causal Bayesian networks, we must modify the definition of $\hBs$
to include the assertion that each nonroot node in $\Bs$ is a direct
causal effect of its parents.  Consequently, the property of
hypothesis equivalence is contradicted by the new definition.
Nonetheless, we have found that the assumption of likelihood
equivalence is reasonable for learning causal networks in many
domains.  (For a detailed discussion of this point, see
Heckerman in this proceedings.\nocite{H95uai})

The next assumption was adopted implicitly in our previous work.

\begin{assumption}[Structure Possibility] 
\label{ass:cposs}
Given a domain $U$, $p(\hBsc|\xi)>0$ for all complete network
structures $\Bsc$.
\end{assumption}

As we shall see, the assumption allows us to make good use of the
property of likelihood equivalence.  Although it is an assumption of
convenience, we have found it to be reasonable for many real-world
network-learning problems.

The remaining two assumptions are abstractions of assumptions made
either explicitly or implicitly by all researchers who have considered
Bayesian-network learning (e.g., Cooper and Herskovits, 1991, 1992;
Buntine, 1991; Spiegelhalter et al.,
1993).\nocite{CH91tr,Cooper92,Buntine91,SDLC93} These assumptions are
made mostly for computational convenience, although they are
reasonable for many domains.

\begin{assumption}[Global Parameter Independence] \label{ass:pi}
For all network structures $\Bs$, 
\[ 
p(\TBs|\hBs,\xi) = \prod_{i=1}^n p(\Theta(x_i,\Pi_i)|\hBs,\xi) 
\] 
\end{assumption} 

\noindent
\Ass{ass:pi} says that the parameters 
associated with each variable in a network structure are independent.
This assumption was first introduced under the name of global
independence by Spiegelhalter and Lauritzen
(1990)\nocite{Spiegelhalter90}.

\begin{assumption}[Parameter Modularity] \label{ass:pm}
Given two network structures $\Bsone$ and $\Bstwo$
such that $p(\hBsone|\xi)>0$ and $p(\hBstwo|\xi)>0$,
if $x_i$ has the same parents in $\Bsone$ and $\Bstwo$, 
then
\[
p(\Theta(x_i,\Pi_i)|\hBsone,\xi) = p(\Theta(x_i,\Pi_i)|\hBstwo,\xi)
\]
\end{assumption}

\noindent
For example, in our two-binary-variable domain, $x$ has the same
parents (none) in the network structure $x \rightarrow y$ and the
structure contains no arc.  Consequently, the probability density for
$\Theta(x,\emptyset)$ would be the same for both of these structures.
We call this property parameter modularity, because it says that the
densities for parameters $\Theta(x_i,\Pi_i)$ depend only on the
structure of the network that is local to variable $x_i$---namely, on
the parents of $x_i$.

Given Assumptions~\ref{ass:le} through \ref{ass:pm}, we can construct
the priors $p(\TBs|\hBs,\xi)$ for every network structure $\Bs$ in
$U$ from the single prior $p(\Th{U}|\hBsc,\xi)$, where $\Bsc$
is any complete network structure for $U$.  As an illustration of this
construction, consider again our two-binary-variable domain.  Given
the prior density
$p(\ta{xy},\ta{x\bar{y}},\ta{\bar{x}y}|\hBxtoy,\xi)$, we construct
the priors $p(\TBs|\hBs,\xi)$ for each of the three network structures
in the domain.  First, consider the network structure $x
\rightarrow y$.  The joint-space parameters and parameters for this 
structure are related as follows:
\[
\ta{xy}=\ta{x}\ta{y|x} \ \ \ \ \ 
\ta{\bar{x}y}=(1-\ta{x})(\ta{y|\bar{x}}) \ \ \ \ \
\ta{x\bar{y}}=\ta{x}(1-\ta{y|x}) 
\]
Thus, we may obtain
$p(\ta{x}, \ta{y|x}, \ta{y|\bar{x}}|\hBxtoy,\xi)$ from the given
density by changing variables:
\begin{equation} \label{eq:j1}
p(\ta{x}, \ta{y|x}, \ta{y|\bar{x}}|\hBxtoy,\xi) =
  J_{x \rightarrow y} 
  \cdot p(\ta{xy}, \ta{\bar{x}y}, \ta{x\bar{y}}|\hBxtoy,\xi)
\end{equation}
where $J_{x \rightarrow y}$ is the Jacobian of the transformation
\begin{eqnarray} \label{eq:j2}
J_{x \rightarrow y} & = &
  \left| \begin{array}{ccc} 
    \dXdY{\ta{xy}}{\ta{x}} 
      & \dXdY{\ta{\bar{x}y}}{\ta{x}}
      & \dXdY{\ta{x\bar{y}}}{\ta{x}} \\
    \dXdY{\ta{xy}}{\ta{y|x}} 
      & \dXdY{\ta{\bar{x}y}}{\ta{y|x}}
      & \dXdY{\ta{x\bar{y}}}{\ta{y|x}} \\
    \dXdY{\ta{xy}}{\ta{y|\bar{x}}} 
      & \dXdY{\ta{\bar{x}y}}{\ta{y|\bar{x}}}
      & \dXdY{\ta{x\bar{y}}}{\ta{y|\bar{x}}}
  \end{array} \right| \nonumber \\
& = & \ta{x}(1-\ta{x})
\end{eqnarray}
The Jacobian $J_{\Bsc}$ for the transformation from $\Th{U}$ to
$\TBsc$ in an arbitrary discrete domain is given in \Sec{sec:bde}.

Next, consider the network structure $x \leftarrow y$.  By 
\Ass{ass:cposs}, the hypothesis $\hBytox$ is also
possible, and, by likelihood equivalence, we have $p(\ta{xy},
\ta{\bar{x}y}, \ta{x\bar{y}}|\hBytox,\xi) = p(\ta{xy},
\ta{\bar{x}y}, \ta{x\bar{y}}|\hBxtoy,\xi)$.
Therefore, we can compute the density for the network structure $x
\leftarrow y$ using the Jacobian $J_{x \leftarrow y} =
\ta{y}(1-\ta{y})$.

\begin{figure}  
\begin{center} 
\leavevmode 
\includegraphics[width=4.0in]{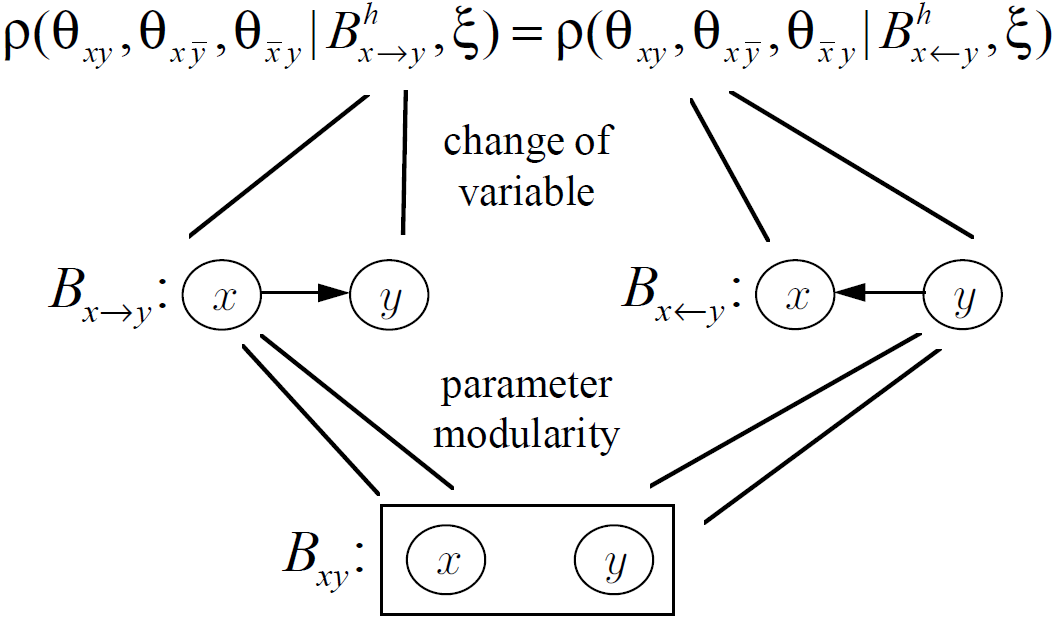}
\end{center} 
\caption{
A computation of the parameter densities for the three network
structures of the two-binary-variable domain $\{x,y\}$.  The approach
computes the densities from
$p(\ta{xy},\ta{x\bar{y}},\ta{\bar{x}y}|\hBxtoy,\xi)$, using likelihood
equivalence, global parameter independence, and parameter modularity.}
\label{fig:prior}
\end{figure}

Finally, consider the empty network structure.  Given the
assumption of global parameter independence, we may obtain the densities
$p(\ta{x}|\hBxy,\xi)$ and $p(\ta{y}|\hBxy,\xi)$ separately.  To obtain
the density for $\ta{x}$, we first extract $p(\ta{x}|\hBxtoy,\xi)$
from the density for the network structure $x
\rightarrow y$. This extraction is straightforward, because, by
global parameter independence, the parameters for $x \rightarrow y$
must be independent.  Then, we use parameter modularity, which says
that $p(\ta{x}|\hBxy,\xi)=p(\ta{x}|\hBxtoy,\xi)$.  To obtain the
density for $\ta{y}$, we extract $p(\ta{y}|\hBytox,\xi)$ from the
density for the network structure $x \leftarrow y$, and again apply
parameter modularity.  The approach is summarized in \Fig{fig:prior}.

In general, we have the following construction.

\begin{theorem} \label{thm:prior}
Given domain $U$ and a probability density $p(\Theta_{U}|\hBsc,\xi)$
where $\Bsc$ is some complete network structure for $U$,
Assumptions~\ref{ass:le} through \ref{ass:pm} determine
$p(\TBs|\hBs,\xi)$ for any network structure $\Bs$ in $U$.
\end{theorem}

We note that our construction assumes that Assumptions~\ref{ass:le}
through \ref{ass:pm} are consistent.  We demonstrate consistency
in \Sec{sec:consis}.

\section{A General Metric for Complete Data} \label{sec:be}

In this section, we derive a general metric
from Assumptions~\ref{ass:rs} through \ref{ass:pm} and the
following additional assumption:

\begin{assumption}[Complete Data] \label{ass:nomiss}
The database is complete.  That is, it contains no missing data.
\end{assumption}

\noindent
We make this assumption only as a computational convenience.  The
reader should recognize that random-sample assumption and the
informative priors developed in \Sec{sec:ip} can be used in
conjunction with well-known statistical techniques to score incomplete
databases as well.  Such techniques include filling in missing data
based on the data that is present
\cite{Titterington76,Spiegelhalter90}, the EM algorithm
\cite{Dempster77}, and Gibbs sampling
\cite{MR94}.

Given our assumptions, we obtain the following lemmas.\footnote{The
proofs are simple and are omitted.}

\begin{lemma}[Posterior Parameter Independence] \label{lem:pi}
Given the random-sample assumption (\Ass{ass:rs}), global parameter
independence (\Ass{ass:pi}), and the assumption of no missing data
(\Ass{ass:nomiss}), we have
\[ 
p(\TBs|\Db,\hBs,\xi) = 
  \prod_{i=1}^n p(\Theta(x_i,\Pi_i)|\Db,\hBs,\xi) 
\] 
for all network structures $\Bs$ ($p(\hBs|\xi)>0$) and databases $\Db$.
\end{lemma}

\begin{lemma}[Posterior Parameter Modularity] \label{lem:pm}
Given the random-sample assumption (Assumption~\ref{ass:rs}), 
global parameter independence (\Ass{ass:pi}), 
parameter modularity (\Ass{ass:pm}), and
the assumption of no missing data (\Ass{ass:nomiss}), if $x_i$ has the
same parents in any two network structures $\Bsone$ and
$\Bstwo$ ($p(\hBsone|\xi)>0, p(\hBstwo|\xi)>0$), then
\[
p(\Theta(x_i,\Pi_i)|\Db,\hBsone,\xi) = p(\Theta(x_i,\Pi_i)|\Db,\hBstwo,\xi)
\]
for all databases $\Db$.
\end{lemma}

In the following lemma and in subsequent discussions, we need the
notion of a {\em database $\Db$ restricted to $X\subseteq U$}---that is
the projection of database $\Db$ onto the subset $X$---denoted $D^X$.
For example, given domain $U=\{x_1,x_2,x_3\}$ and database $D =
\{C_1=\{x_1=1,x_2=2,x_3=1\}, C_2=\{x_1=2,x_2=2,x_3=1\}\}$, we have
$D^{\{x_1,x_2\}} = \{C_1 = \{x_1=1,x_2=2\}, C_2=\{x_1=2,x_2=2\}\}$.

\begin{lemma} \label{lem:ignore}
Let $X$ be a subset of $U$, and $\Bsc$ ($p(\hBsc|\xi)>0$) be a
complete network structure for any ordering where the variables in $X$
come first.  Given the random-sample assumption
(Assumption~\ref{ass:rs}), global parameter independence
(\Ass{ass:pi}), and the assumption of no missing data
(\Ass{ass:nomiss}),
\[
p(X|\Db,\hBsc,\xi) = p(X|\Db^X,\hBsc,\xi)
\]
for all databases $\Db$.
\end{lemma}

\noindent
Readers familiar with the concept of d-separation will recognize that
Lemmas~\ref{lem:pi} and \ref{lem:ignore} can be readily obtained from
graphical manipulations applied to the Bayesian-network representation
of the random-sample assumption and the assumption of global parameter
independence.

We can now derive the general metric.

\begin{theorem} \label{thm:be}
Given a domain $U$, let $\Bs$ be any network structure for $U$
and $\Bsc$ be a some complete network structure for $U$.
Then, given Assumptions~\ref{ass:le} through \ref{ass:nomiss},
\begin{equation} \label{eq:be}
p(\Db,\hBs|\xi) = p(\hBs|\xi) \cdot \prod_{i=1}^n 
  \frac{p(\Db^{\Pi_i,x_i}|\hBsc,\xi)}{
        p(\Db^{\Pi_i}|\hBsc,\xi)}
\end{equation}
for any database $\Db$.
\end{theorem}

\noindent {\bf Proof:}
From the rules of probability, we obtain
\begin{eqnarray*} \label{eq:be1}
\lefteqn{ p(\Db|\hBs,\xi) = \prod_{l=1}^m \int p(\TBs|\Db_l,\hBs,\xi) } \\
&& \cdot
  \prod_{i=1}^n p(x_{il}|x_{1l},\ldots,x_{(i-1)l},\Db_l,\TBs,\hBs,\xi)
  \ d\TBs
\end{eqnarray*}
For every $x_i$ with parents $\Pi_i$ in $\Bs$, let
$B_{SC,\Pi_i,x_i}$ be a complete network structure with variable
ordering $\Pi_i$, $x_i$ followed by the remaining variables.  
By \Ass{ass:cposs}, $p(B_{SC,\Pi_i,x_i}|\xi) > 0$.
Using \Ass{ass:rs} and Lemmas~\ref{lem:pi}
and \ref{lem:pm}, we get
\begin{eqnarray*} \label{eq:be2}
\lefteqn{ p(\Db|\hBs,\xi) = \prod_{l=1}^m \int \prod_{i=1}^n 
  p(\Theta(x_i,\Pi_i)|\Db_l,B_{SC,\Pi_i,x_i},\xi) } \\
&& \cdot
  p(x_{il}|\Pi_{1l},\Db_l,\Theta(x_i,\Pi_i),B_{SC,\Pi_i,x_i},\xi) 
  \ d\TBs
\end{eqnarray*}
Decomposing the integral over $\TBs$ into integrals over the
individual parameter sets $\Theta(x_i,\Pi_i)$, and 
performing the integrations, we have
\begin{displaymath} \label{eq:be3}
p(\Db|\hBs,\xi) = \prod_{l=1}^m \prod_{i=1}^n 
  p(x_{il}|\Pi_{1l},\Db_l,B_{SC,\Pi_i,x_i},\xi)
\end{displaymath}
Also, using \Lem{lem:ignore}, we obtain
\begin{eqnarray} \label{eq:be4}
p(\Db|\hBs,\xi) & = & \prod_{l=1}^m \prod_{i=1}^n 
  \frac{p(x_{il},\Pi_{1l}|\Db_l,B_{SC,\Pi_i,x_i},\xi)}{
        p(\Pi_{1l}|\Db_l,B_{SC,\Pi_i,x_i},\xi)} \nonumber \\*[9pt]
& = & \prod_{l=1}^m \prod_{i=1}^n 
  \frac{p(x_{il},\Pi_{1l}|\Db_l^{\Pi_i,x_i},B_{SC,\Pi_i,x_i},\xi)}{
  p(\Pi_{1l}|\Db_l^{\Pi_i},B_{SC,\Pi_i,x_i},\xi)} \nonumber \\*[9pt]
& = & \prod_{i=1}^n 
  \frac{p(\Db^{\Pi_i,x_i}|B_{SC,\Pi_i,x_i},\xi)}{
  p(\Db^{\Pi_i}|B_{SC,\Pi_i,x_i},\xi)}
\end{eqnarray}
By likelihood equivalence, we have that $p(\Db|B_{SC,\Pi_i,x_i},\xi) =
p(\Db|\hBsc,\xi)$.  Consequently, for any subset $X$ of $U$, we obtain
$p(\Db^X|B_{SC,\Pi_i,x_i},\xi) = p(\Db^X|\hBsc,\xi)$ by summing over the
variables in $D^{U \setminus X}$.  Applying this result to
\Eq{eq:be4}, we get \Eq{eq:be}. \qed

We call \Eq{eq:be} the Be ($B$ayesian likelihood $e$quivlent) metric.

\section{Special-Case Metrics}

Our general metric is powerful, because it tells us that if we know how
to compute $p(\Db^X|\hBsc,\xi)$ for any subset $X$ of $U$ under the
assumption that the domain contains no structure (i.e., there are no
independencies), then we can compute the probability of any database
when there is structure.  Therefore, the Be metric allows us to
leverage much of the work in the statistics literature, as
statisticians have long dealt with the former problem.  In this
section, we illustrate this claim by deriving likelihood-equivalent
metrics for the discrete and Gaussian cases.

\subsection{The BDe Metric} \label{sec:bde}

Suppose all variables in $U$ are discrete.  Recall that we use
$\ta{X=k_X|Y=k_Y}$ denote the multinomial parameter
corresponding to probability $p(X=k_X|Y=k_Y,\xi)$.  In
addition, we use $\Th{X|Y}$ denote the collection of parameters
$\ta{X=k_X|Y=k_Y}$ for all states of sets $X$ and $Y$.
If $Y$ is empty, we simply write $\Th{X}$.  Thus, for example,
$\Th{U}=\Th{x_1,\ldots,x_n}$ represents the multinomial parameters of
the joint space of $U$.

Let us assume that the parameter set
$\Theta_U$ has a Dirichlet distribution when
conditioned on a hypothesis corresponding to some complete network
structure $\Bsc$:
\begin{equation} \label{eq:joint-dir}
p(\Theta_{x_1,\ldots,x_n}|\hBsc) =
  \prod_{x_1,\ldots,x_n}  
    \theta_{x_1,\ldots,x_n}^{
      N'_{\Bsc} p(x_1,\ldots,x_n|\hBsc,\xi) - 1}
\end{equation}
where $N'_{\Bsc}$ is the \textcolor{blue}{effective} sample size of the Dirichlet
distribution associated with a complete network structure $\Bsc$.
DeGroot (1970, p.\ 50)\nocite{DeGroot70} shows that, 
for any subset $X$ of $U$, $\Theta_X$ also has a Dirichlet distribution:
\begin{equation} \label{eq:joint-dirX}
p(\Theta_X|\hBsc,\xi) =
  \prod_{X}  
    \theta_{X}^{
      N'_{\Bsc} p(X|\hBsc,\xi)-1}
\end{equation}
Now, it is a well-known statistical result that, if a discrete variable
$x$ with $r$ states has a Dirichlet distribution with exponents
$N'_1-1,\ldots,N'_{r}-1$, then
\begin{equation} \label{eq:dir-ss}
p(\Db|\xi) = 
  \frac{\Gamma(\sum_{k=1}^r N'_k)}{\Gamma(\sum_{k=1}^r N'_k + N_k)} \ 
  \prod_{k=1}^r \frac{\Gamma(N'_k+N_k)}{\Gamma(N'_k)}
\end{equation}
where $\Db$ is a database for variable $x$ and $N_k$ is the number of
times $x$ takes on state $k$ in $\Db$.  Also, because $U$ is discrete,
any subset $X$ of $U$ can also be thought of as a single discrete
variable with $\prod_{x_i \in X} r_i$ states.  Therefore,
\Eqs{eq:joint-dirX} and \ref{eq:dir-ss} allow us to compute each term
in the Be metric (\Eq{eq:be}).  To express the resulting metric for a
given network structure $\Bs$, we use $q_i = \prod_{x_i \in \Pi_i}
r_i$ to denote the number of states of $\Pi_i$ in $\Bs$, and $\Pi_i=j$
to denote that $\Pi_i$ has assumed the $j$th state,
$j=1,\ldots,q_i$.

\begin{theorem}[BDe Metric] \label{thm:bde}
Given domain $U$, and network structure $\Bs$ and database $\Db$ for
$U$, let $N_{ijk}$ denote the number of times that $x_i=k$ and
$\Pi_i=j$ in the database $\Db$; and let $N_{ij} = \sum_{k=1}^{r_i}$
denote the number of times that $\Pi_i=j$ in a database $\Db$.  Then, if
$p(\Th{U}|\hBsc,\xi)$ is Dirichlet with 
\textcolor{blue}{effective} sample size $N'$
for some complete network structure $\Bsc$, and if
Assumptions~\ref{ass:le} through \ref{ass:nomiss} hold, then
\begin{eqnarray} \label{eq:bde}
p(\Db,\hBs|\xi) & = & p(\hBs|\xi) \cdot \prod_{i=1}^n \prod_{j=1}^{q_i}
  \frac{\Gamma(N'_{ij})}{\Gamma(N'_{ij}+N_{ij})} \nonumber \\
&& \cdot
  \prod_{k=1}^{r_i} \frac{\Gamma(N'_{ijk} + N_{ijk})}{\Gamma(N'_{ijk})}  
\end{eqnarray}
where
\[
\Npijk = N' \cdot p(x_i=k,\Pi_i=j|\hBsc,\xi)
\]
\begin{equation} \label{eq:npijk}
\Npij = \sum_{k=1}^{r_i} \Npijk = N' \cdot p(\Pi_i=j|\hBsc,\xi)
\end{equation}
\end{theorem}

\Eqs{eq:bde} and \ref{eq:npijk} are the BDe ($B$ayesian
$D$irichlet likelihood $e$quivalent) metric, originally derived in
Heckerman et al. (1994).  

The assumption that $p(\Theta_U|\hBsc,\xi)$ is Dirichlet is not as
arbitrary as it may seem at first glance.  In discrete domains, we can
assume not only that the parameters corresponding to each variable are
independent, but that the parameters corresponding to each state of
every variable's parents are independent.  Spiegelhalter and Lauritzen
(1990) call this added assumption {\em local independence}.  Geiger
and Heckerman (in this proceedings)\nocite{GH95uai} show that
likelihood equivalence, structure possibility, global and local
parameter independence, and the assumption that
$p(\Theta_U|\hBsc,\xi)$ is positive imply that $p(\Theta_U|\hBsc,\xi)$
must be Dirichlet.

\subsection{The BGe Metric} \label{sec:bge}

Suppose that all variables in $U=\x$ are continuous, and that the
database is a random sample from a multivariate-normal distribution.
Let us assume that the parameter set $\{\vecmu,\matW\}$ has a
normal--Wishart distribution when conditioned on $\hBsc$ for some
complete network structure $\Bsc$.  Namely, assume that
$p(\vecmu|\matW,\hBsc,\xi)$ is a multivariate-normal distribution with
mean $\vecmu_0$ and precision matrix $N'_{\mu}
\matW$ ($\Npmu > 0$); and that $p(\matW|\hBsc,\xi)$ is a Wishart
distribution with $\Nptau$ degrees of freedom $(\Nptau > n-1)$ and
positive-definite matrix $\matTau_0$.  That is,
\begin{equation} \label{eq:w}
p(\matW|\hBsc,\xi) =  
  c \ |\matW|^{(\Nptau-n-1)/2} e^{-1/2\tr\{\matTau_0 \matW\}}  
\end{equation}
where $c$ is a normalization constant [DeGroot, 1970, p.\ 57].  

It is well known that the normal--Wishart distribution is a conjugate
family for multivariate-normal sampling (e.g., DeGroot, 1970,
p.\ 178).\nocite{DeGroot70} Given a database
$D=\{\vec{x_1},\ldots,\vec{x_m}\}$, let $\xvecbar_m$ and $\matS_m$
denote its sample mean and 
\textcolor{red}{scatter matrix}, 
respectively.  Then, given the
normal--Wishart prior we have described, the posterior density
$p(\vecmu,\matW|\Db,\hBsc,\xi)$ is also a normal--Wishart distribution.
In particular, $p(\vecmu|\matW,\Db,\hBsc,\xi)$ is multivariate normal
with mean vector $\vecmu_m$ given by
\begin{equation} 
\label{eq:updatemeans} 
\vecmu_m = \frac{\Npmu \vecmu_0 + m \xvecbar_m}{\Npmu+m}
\end{equation} 
and precision matrix $(\Npmu+m)\matW$; and 
$p(\matW|\Db,\hBsc,\xi)$ is a Wishart distribution with
$\Nptau+m$ degrees of freedom and matrix $\matTau_m$ given
by
\begin{equation} \label{eq:updateTau} 
\matTau_m =  
\matTau_0 + \matS_m + \frac{\Npmu m}{\Npmu+m}
  (\vecmu_0-\xvecbar_m)(\vecmu_0-\xvecbar_m)' 
\end{equation} 
From these equations, we see that $\Npmu$ and $\Nptau$ can be thought
of as \textcolor{blue}{effective sample sizes for the 
normal and Wishart components
of the prior, respectively.}

Given domain $U=\{x_1,\ldots,x_n\}$, subset $X$ of $U$ with $l$ elements, 
and vector
$\vec{y}=(y_1,\ldots,y_n)$, let $\vec{y}^X$ denote the vector formed
by the components $y_i$ of $\vec{y}$ such that $x_i \in X$.
Similarly, given matrix $\matM$, let $\matM^X$ denote the submatrix of
$\matM$ containing elements $m_{ij}$ such that $x_i,x_j \in X$.  It is
well known that if $\Db$ is a random sample from an $n$-dimensional
multivariate-normal distribution whose parameters $\{\vecmu,\matW\}$
have a normal--Wishart distribution with constants $\vecmu_0$, $\Npmu$,
$\matTau_0$, and $\Nptau$, then $D^X$ is a random sample from an
$|X|$-dimensional multivariate distribution with parameters
$\{\vecmu^X,\matW^X\}$, and these parameters have normal--Wishart
distribution with constants $\vecmu_0^X$, $\Npmu$, $\matTau_0^X$, and
$\Nptau\textcolor{red}{-n+l}$.  
Furthermore, the formula for $p(\Db|\hBsc,\xi)$ given the
normal--Wishart prior is known \textcolor{blue}{(Geiger and Heckerman, 1994)}.
Consequently, the evaluation of
$p(\Db^X|\hBsc,\xi)$ in \Eq{eq:be} is straightforward.

\begin{theorem}[BGe Metric] \label{thm:bge}
\begin{sloppypar}
Given domain $\x=\{x_1,\ldots,x_n\}$, assume
$p(\vecmu,\matW|\hBsc,\xi)$ is an $n$-dimensional normal--Wishart
distribution with constants $\vecmu_0, \Npmu,
\matTau_0,$ and $\Nptau$.  Given a database $D=\{C_1,\ldots,C_m\}$
and a subset $X$ of $\x$ with $l$ elements,
Assumptions~\ref{ass:le} through
\ref{ass:nomiss} imply the Be metric, where each term is given by
\end{sloppypar}
\begin{eqnarray} \label{eq:bge}
p(\Db^X|\hBsc,\xi) & = &
  \pi^{-lm/2} \left(\frac{\Npmu}{\Npmu+m}\right)^{l/2} \\
&& \cdot
  \frac{c(l,m+\Nptau\textcolor{red}{-n+l})}{c(l,\Nptau\textcolor{red}{-n+l})} 
  |\matTau_0^X|^{\frac{\Nptau\textcolor{red}{-n+l}}{2}} |\matTau_m^X|^{-\frac{m+\Nptau\textcolor{red}{-n+l}}{2}} 
  \nonumber
\end{eqnarray}
where
\begin{equation} \label{eq:cln} 
c(l,\Nptau) = \prod_{i=1}^l \Gamma\left(\frac{\Nptau+1-i}{2}\right)
\end{equation} 
and $\matTau_m$ is the matrix of the posterior
normal--Wishart distribution given by \Eq{eq:updateTau}.
\end{theorem}

The Be metric in combination with \Eq{eq:bge} defines the BGe
($B$ayesian $G$aussian likelihood $e$quivalent) metric, originally
derived in Geiger and Heckerman (1994).  We note that assumptions
similar to those used to show the inevitability of the Dirichlet
distribution for discrete domains imply that the normal-Wishart
assumption is inevitable for Gaussian domains (see Geiger and
Heckerman in this proceedings).

The BDe and BGe metrics may be combined to score domains containing
both discrete variables and continuous variables.  Namely, let $U=U_d
\cup U_c$ where all variables in $U_d$ and $U_c$ are discrete and
continuous, respectively.  Suppose that the observations of $U_d$ in
the database are a random sample from a multivariate-discrete
distribution, and the observations of the $U_c$ given each state of
$U_d$ are a random sample from a multivariate-normal distribution.
Finally, suppose that $\Th{U_d}$ has a Dirichlet distribution, and
that $\Th{U_c|U_d=k}$ has a normal--Wishart distribution for every
state $k$ of $U_d$.  Then, we can apply the Be metric to any network
structure $\Bs$ where the variables in $U_d$ precede the variables in
$U_c$, using
\Eq{eq:dir-ss} to evaluate terms for discrete variables, and
\Eqs{eq:bge} and \ref{eq:cln} to evaluate terms for continuous
variables.

\section{Informative Priors from a Prior Network} \label{sec:prior}

Given our assumptions, $p(\Th{U}|\hBsc,\xi)$ determines a Bayesian
scoring metric.  In this section, we discuss the assessment of this
distribution.

For discrete domains, we can assess $p(\Th{U}|\hBsc,\xi)$ by assessing
(1) the joint probability distribution for the first cases to be seen
in the database $p(U|\hBs,\xi)$ and (2) the 
\textcolor{blue}{effective} sample size
$N'$ for the domain.  Methods for assessing $N'$ are discussed in
(e.g.) Heckerman et al. (1995).\nocite{HGC95ml} To assess
$p(U|\hBs,\xi)$, we can construct a Bayesian network for the first
case to be seen.  We call this Bayesian network a {\em prior network.}
The unusual aspect of this assessment is the conditioning hypothesis
$\hBsc$ (see Heckerman et al. [1995] for a discussion).

We can assess $p(\Th{U}|\hBsc,\xi)$ in the Gaussian case using a
prior network as well.  In this case, however, we require two
\textcolor{blue}{effective}
samples sizes ($\Npmu>0$ and $\Nptau>n-1$).  The details
are discussed in last year's proceedings \cite{GH94uai}.  Examples of
the assessment of $p(\Th{U}|\hBsc,\xi)$ for discrete and Gaussian
domains, and examples of the metrics that result from these
assessments are also given in last year's proceedings.

\section{Consistency of the Assumptions} \label{sec:consis}

The assumptions of likelihood equivalence, structure possibility,
global parameter independence, and parameter modularity may not be
consistent.  In particular, the assumptions of global parameter
independence and modularity are constraints on parameter densities
among individual network structures, whereas likelihood equivalence is
a constraint on parameter densities among network-structure
equivalence classes.  Furthermore, our choices $p(\Th{U}|\hBsc,\xi)$
is Dirichlet and $p(\vec{\mu},W|\hBsc,\xi)$ is normal-Wishart may not
be consistent with the assumptions of likelihood equivalence and
global parameter independence.  In this section, we demonstrate
consistency in each case.

\subsection{Consistency of the Dirichlet Assumption} \label{sec:bde-consis}

First, we show that the assumption $p(\Th{U}|\hBsc,\xi)$ is Dirichlet
is consistent with the assumptions of likelihood equivalence and
global parameter independence for complete network structures.

To see the potential for inconsistency, consider again our approach
for constructing priors in the two-binary-variable domain.  Suppose we
choose the density
\[
p(\ta{xy}, \ta{\bar{x}y}, \ta{x\bar{y}}|\hBxtoy,\xi) =
  \frac{c}{(\ta{xy} + \ta{x\bar{y}})  (1- (\ta{xy} + \ta{x\bar{y}}))} 
\]
\vspace{-\medskipamount}
\[
\ \ \  = \frac{c}{\ta{x} (1 - \ta{x})}
\]
where $c$ is a normalization constant.  By \Eqs{eq:j1} and \ref{eq:j2}
we obtain 
\[
p(\ta{x}, \ta{y|x}, \ta{y|\bar{x}}|\hBxtoy,\xi) = c
\]
for the network structure $x \rightarrow y$.  This density satisfies
the assumption of global (and local) parameter independence.  Using
likelihood equivalence, however, we have for the network structure
$\ytox$
\[
p(\ta{y}, \ta{x|y}, \ta{x|\bar{y}}|\hBytox,\xi) = 
  \frac{c \cdot \ta{y}(1-\ta{y})}{\ta{x} (1 - \ta{x})} = 
\]
\vspace{-\medskipamount}
\[
  \frac{c \cdot \ta{y}(1-\ta{y})}{
    (\ta{y}\ta{x|y}+(1-\ta{y})\ta{x|\bar{y}})
    (1 - (\ta{y}\ta{x|y}+(1-\ta{y})\ta{x|\bar{y}}))}
\]
This density satisfies neither global (nor local) parameter
independence.

When $p(\Th{U}|\hBsc,\xi)$ is Dirichlet, however, likelihood
equivalence implies global (and local) parameter independence for all
complete network structures.  This result is proved for the
two-variable case in Dawid and Lauritzen (1993, Lemma 7.2) and for the
general case in Heckerman et al. (1995, Theorem 3), which we summarize
here.

\begin{theorem} \label{thm:jac}
Let $\Bsc$ be any complete network structure for domain
$U=\{x_1,\ldots,x_n\}$.  The Jacobian for the transformation from
$\Theta_{U}$ to $\TBsc$ is
\begin{equation} \label{eq:jac}
J_{\Bsc} = \prod_{i=1}^{n-1} \prod_{x_1,\ldots,x_{i}} [
\ta{x_i|x_1,\ldots,x_{i-1}} ]^{[\prod_{j=i+1}^n r_j] - 1}
\end{equation}
\end{theorem}

\begin{theorem} \label{thm:consis}
Given a domain $U = \{x_1,\ldots,x_n\}$, if the
parameters $\Theta_U$ have a Dirichlet distribution with
parameters $N'_{x_1,\ldots,x_n}$---that is,
\begin{equation} \label{eq:given}
p(\Theta_{U}|\xi) = 
c \cdot \prod_{x_1,\ldots,x_n} 
  [\theta_{x_1,\ldots, x_n}]^{N'_{x_1,\ldots,x_n}-1}
\end{equation} 
then, for any complete network structure $\Bsc$ in $U$, the density
$p(\TBsc|\xi)$ satisfies global and local parameter independence.  In
particular,
\begin{equation} \label{eq:phidir3a}
p(\TBsc|\xi)= 
c \cdot \prod_{i=1}^n \prod_{x_1,\ldots, x_i}
[\theta_{x_i | x_1,\ldots, x_{i-1}}]^ {N'_{x_i|x_1,\ldots,x_{i-1}}-1}
\end{equation} 
where
\begin{equation} \label{eq:phidir3b}
N'_{x_i|x_1,\ldots,x_{i-1}} = 
  \sum_{x_{i+1},\ldots,x_n} N'(x_1,\ldots,x_n)
\end{equation} 
\end{theorem}

\noindent {\bf Proof:}  The result follows by 
multiplying the right-hand-side of \Eq{eq:given} by the
Jacobian in \Thm{thm:jac}, using the relation
$\theta_{x_1,\ldots,x_n}= \prod_{i=1}^n
\theta_{x_i| x_1,\ldots,x_{i-1}}$, and collecting
powers of $\ta{x_i|x_1,\ldots,x_{i-1}}$. \qed

It is interesting to note that each set of conditional parameters
$\Th{x_i|x_1,\ldots,x_{i-1}}$ also has a Dirichlet distribution.

\subsection{Consistency of the Normal--Wishart Assumption} 
\label{sec:bge-consis}

Next, we show that the assumption $p(\vecmu,\matW|\hBsc,\xi)$ is
normal--Wishart is consistent with the assumptions of likelihood
equivalence and global parameter independence for complete network
structures.

\begin{theorem} \label{thm:jac-w} 
The Jacobian for the change of variables from $W$ to
$\{\vecv,\matB\}$ is given by 
\begin{equation}  \label{eq:jac-w} 
J_{\vecv,\matB} = | \dXdY{\matW}{\vecv \matB} | =  
  \prod_{i=1}^n v_i^{-(i+1)} 
\end{equation} 
\end{theorem} 
 
\noindent {\bf Proof:} Let $J(i)$ denote the Jacobian for the first $i$ 
variables in $\matW$.  Then $J(i)$ has the following form: 
\begin{equation} 
\label{eq:G-jacmatrix} 
J(i) = 
\left| \begin{array}{ccc} 
J(i-1)  & 0 & 0  \\ 
0       & -\frac{1}{v_i} I_{i-1,i-1} & 0  \\ 
0       &  0          & -\frac{1}{v_i^2} 
\end{array} \right| 
\end{equation} 
where $I_{k,k}$ is the identity matrix of size $k \times k$. 
Thus, we have
\begin{equation} 
\label{eq:G-jacobian} 
J(i) = \frac{1}{v_i^{i+1}} \cdot J(i-1)
\end{equation} 
which gives \Eq{eq:jac-w}. \qed 
 
\begin{theorem} \label{thm:jac-m} 
The Jacobian for the change of variables from $\vecmu$ to $\vecm$
is given by $J_{\vecm} = 1$.
\end{theorem} 

\noindent {\bf Proof:}  From \Eq{eq:mi}, $J_{\vecm}$ is the determinant
of a triangular matrix whose diagonal elements are 1. \qed

\begin{theorem} \label{thm:pinw} 
If $\{\vecmu,\matW\}$ has a normal--Wishart distribution
given background information $\xi$, then
\[ 
p(\vecm, \vecv, \matB|\xi) = \prod_{i=1}^n p(m_i, v_i, \vecbi|\xi) 
\] 
\end{theorem} 
 
\noindent {\bf Proof:}  To prove the theorem, we factor
$p(\vecm|\vecv,\matB,\xi)$ and $p(\vecv,\matB|\xi)$ separately.
By assumption, we know that $p(\vecmu|W)$ is a multivariate-normal
distribution with mean $\mu_0$ and precision matrix $\Npmu W$.
Transforming this result to conditional distributions for $\mu_i$,
we obtain
\begin{eqnarray*}
\lefteqn{ p(\mu_i|\mu_1,\ldots,\mu_{i-1},\vecv,\matB,\xi) = 
  \left(\frac{\Npmu}{2\pi v_i}\right)^{1/2} } \\
&& \cdot \exp \left\{ 
    \frac{\left(\mu_i - \mu_{0i} - \sum_{j=1}^{i-1} b_{ji} 
      (\mu_j - \mu_{0j}) \right)^2}{2v_i/\Npmu}
  \right\}
\end{eqnarray*}
for $i=1,\ldots,n$.
Letting $m_{0i} = \mu_{0i} - \sum_{j=1}^{i-1} b_{ji} \mu_{0j}$ for
each $i$, we get
\begin{eqnarray*}
\lefteqn{ p(\mu_i|\mu_1,\ldots,\mu_{i-1},\vecv,\matB,\xi) = 
  \left(\frac{\Npmu}{2\pi v_i}\right)^{1/2} } \\
&& \cdot \exp \left\{ 
    \frac{(m_i - m_{0i})^2}{2v_i/\Npmu}
  \right\} \ \ \ \ \ \ \ \ \ \ \ \ \ \ \ \ \ \ \ \ \ \ \ 
\end{eqnarray*}
Thus, collecting terms for each $i$ and using the
Jacobian $J_{\vecm}=1$, we have 
\begin{equation} \label{eq:pinw1}
p(\vecm|\vecv,\matB,\xi) = \prod_{i=1}^n n(m_{0i},\Npmu/v_i)
\end{equation}
In addition, by assumption, we have
\begin{equation} \label{eq:p-ind-w1}
p(\matW|\xi) =  
  c |\matW|^{(\alpha-n-1)/2} e^{-1/2\tr\{\matTau_0 \matW\}}  
\end{equation}
From \Eq{eq:shachter}, we have
\[ 
|\matW(i)| = \frac{1}{v_i}|\matW(i-1)| = \prod_{i=1}^n v_i^{-1} 
\] 
so that the determinant in \Eq{eq:p-ind-w1} factors as a function of 
$i$.  Also, \Eq{eq:shachter} implies (by induction) that each element 
$w_{ij}$ in $\matW$ is a sum of terms each being a function of $\vecbi$ 
and $v_i$. Consequently, the exponent in \Eq{eq:p-ind-w1} factors as a 
function of $i$.  Thus, given the Jacobian $J_{\vecv,\matB}$, which
also factors as a function of $i$, we obtain
\begin{equation} \label{eq:pinw2}
p(\vecv,\matB|\xi) = \prod_{i=1}^n p(v_i, \vecb_i|\xi)
\end{equation}
\Eqs{eq:pinw1} and \ref{eq:pinw2} imply the theorem. \qed

\subsection{Consistency of Likelihood Equivalence, Structure Possibility,
Parameter Independence, and Parameter Modularity}

As mentioned, the assumptions of likelihood equivalence, structure
possibility, global parameter independence, and parameter modularity
may not be consistent.  To understand the potential for inconsistency,
note that we obtained the Be metric (\Eq{eq:be}) for all network
structures using likelihood equivalence applied only to complete
network structures in combination with the assumptions of structure
possibility, global parameter independence, parameter modularity.
Thus, it could be that the Be metric for incomplete network structures
is not likelihood equivalent.  Nonetheless, the following theorem
shows that the Be metric is likelihood equivalent for all network
structures---that is, given structure possibility, global parameter
independence, and parameter modularity, likelihood equivalence for
incomplete structures is implied by likelihood equivalence for
complete network structures.  Consequently, the assumptions are
consistent.

\begin{theorem}[Likelihood Equivalence]
If $\Bsone$ and $\Bstwo$ are equivalent network structures for
domain $U$, then, for all databases $\Db$, $p(\Db|\hBsone,\xi) =
p(\Db|\hBstwo,\xi)$, where each likelihood is computed by the Be metric
(\Eq{eq:be}).
\end{theorem}

\noindent {\bf Proof:}
By \Thm{thm:max}, we know that a network structure can be transformed
into an equivalent structure by a series of arc reversals.  Thus, we
can demonstrate likelihood equivalence in general if we can do so for
the case where two equivalent structures differ by a single arc
reversal.  So, let $\Bsone$ and $\Bstwo$ be two equivalent network
structures that differ only in the direction of the arc between $x_i$
and $x_j$ (say $x_i \rightarrow x_j$ in $\Bsone$).  Let $R$ be the set
of parents of $x_i$ in $\Bsone$.  By \Thm{thm:max}, we know that $R
\cup \{x_i\}$ is the set of parents of $x_j$ in $\Bsone$, $R$ is the
set of parents of $x_j$ in $\Bstwo$, and $R \cup \{x_j\}$ is the set
of parents of $x_i$ in $\Bstwo$.  Because the two structures differ
only in the reversal of a single arc, the only terms in the product of
\Eq{eq:be} that can differ are those involving $x_i$ and $x_j$.  For
$\Bsone$, these terms are
\[
\frac{p(\Db^{x_i R}|\hBsc,\xi)}{p(\Db^{R}|\hBsc,\xi)}
  \frac{p(\Db^{x_i x_j R}|\hBsc,\xi)}{p(\Db^{x_i R}|\hBsc,\xi)}
     = \frac{p(\Db^{x_i x_j R}|\hBsc,\xi)}{p(\Db^{R}|\hBsc,\xi)}
\]
whereas for $\Bstwo$, they are
\[
\frac{p(\Db^{x_j R}|\hBsc,\xi)}{p(\Db^{R}|\hBsc,\xi)}
  \frac{p(\Db^{x_i x_j R}|\hBsc,\xi)}{p(\Db^{x_j R}|\hBsc,\xi)}
     = \frac{p(\Db^{x_i x_j R}|\hBsc,\xi)}{p(\Db^{R}|\hBsc,\xi)}
\]
These terms are equal, and consequently, so are the likelihoods. \qed

\section*{Acknowledgments}

We thank Peter Spirtes for identifying an error with \Eq{eq:bge}.

\bibliographystyle{apalike}
%\bibliography{David}

\end{document}